\pdfoutput=1
\documentclass[10pt,twocolumn,twoside,journal]{IEEEtran}

\newcommand{\comment}[1]{}

\usepackage{amsmath,amssymb} 
\usepackage{color}

\usepackage{cs}
\usepackage{mathsymb}
\usepackage{verbatim}
\usepackage{subfigure}
\usepackage{url}

\usepackage{graphicx}
\graphicspath{{./}{./Fig/}{./Figs/}{./Fig/eps/}{./Fig/pdf/}}

\usepackage{algorithm2e}
\let\savedalgorithm\algorithm
\let\savedendalgorithm\endalgorithm
\usepackage{algorithm}
\usepackage{amsmath}
\usepackage{url}
\usepackage{multirow}

\def\Integer{\mathbb{Z}^+}

\begin{document}

\title{Face Detection with Effective Feature Extraction}

\author{
         Sakrapee Paisitkriangkrai,
         Chunhua Shen,
         Jian Zhang
% \thanks
% {
% Copyright (c) 2010 IEEE. Personal use of this material is permitted.
% However, permission to use this material for any other purposes must be obtained
% from the IEEE by sending a request to pubs-permission@ieee.org.
% }
\thanks
{
S. Paisitkriangkrai is with University of Adelaide, SA 5005, Australia.
}
\thanks
{
C. Shen is with NICTA, Canberra Research Laboratory,
Canberra, ACT 2601, Australia,
and also with the Australian National University, Canberra,
ACT 0200, Australia. 
Correspondence should be addressed to C. Shen
(e-mail: chunhua.shen@nicta.com.au). 
}
\thanks
{ 
J. Zhang is with NICTA, Neville Roach Laboratory, Sydney, NSW 1466, Australia.
}
\thanks{
%
% Manuscript received XX, 201X; accepted XX, 201X.
%
NICTA is funded by the Australian Government as represented by
the Department of Broadband, Communications and the Digital
Economy and the Australian Research Council through the ICT
Center of Excellence program. 
This research was also supported in part by the Australian Research
Council through its special research initiative in bionic vision science and
technology grant to Bionic Vision Australia.
}
}

\markboth{October 2010}{}

\maketitle

%===========================================================
\begin{abstract}

    There is an abundant literature on face detection due to its important role in many vision
    applications.  Since Viola and Jones proposed the first real-time AdaBoost based face detector,
    Haar-like features have been adopted as the method of choice for frontal face detection.  In
    this work, we show that simple features other than Haar-like features can also be applied for
    training an effective face detector.  Since, single feature is not discriminative enough to
    separate faces from difficult non-faces, we further improve the generalization performance of
    our simple features by introducing feature co-occurrences.  We demonstrate that our proposed
    features yield a performance improvement compared to Haar-like features.  In addition, our
    findings indicate that features play a crucial role in the ability of the system to generalize.

\end{abstract}

\begin{IEEEkeywords}
    Face detection, boosting, 
    Haar features, 
    histogram of oriented gradients,
    feature co-occurrence.
\end{IEEEkeywords}

\section{Introduction}

  Face detection is an important first step
  for several computer vision applications.
  It was not until recently that face detection problem received considerable attention
  among researchers owing to the impressive performance of
  Viola and Jones' face detector \cite{Viola2004Robust}.
  Their detector was the first algorithm that achieved real-time detection speed and
  high accuracy comparable to previous state of the art methods.
  Their work consists of three contributions. The first contribution is a cascade of classifiers.
  The second contribution is the boosted classifier where a combination of
  linear classifiers is formed to achieve fast calculation time with high accuracy.
  The last contribution is a simple rectangular Haar-like feature which can be extracted
  and computed in fewer than ten Central Processing Unit (CPU) operations using integral image.

  Haar-like wavelet features are defined as a difference between the accumulated
  intensities of filled rectangles and unfilled rectangles.
  %The classical layouts of the Haar rectangle features are illustrated
  %in Fig.~\ref{fig:feat_haar}.
  Several researchers have proposed various approaches to extend the
  robustness and discriminative power of Haar-like features
  \cite{Lienhart2002Extended,Li2004Float}.
  Lienhart \etal proposed a novel set
  of rotated Haar-like features which can also be calculated efficiently
  \cite{Lienhart2002Extended}.
  Li and Zhang later proposed a simple Haar wavelet, which separates Haar-like
  rectangles at some distance apart \cite{Li2004Float}.
  The authors tested their proposed features on multi-view faces and demonstrated
  excellent performance.
  Huang \etal \cite{Huang2007High} further extended Haar-like features in a slightly
  different way.
  Instead of using rectangles, they proposed sparse granular features, which represent
  a sum of pixel intensities in a square.
  An efficient weak learning algorithm is introduced which adopts heuristic search method
  in pursuit of discriminative sparse granular features.
  Since, sparse granular features have a smaller rectangular region than Haar-like
  features; it has a better discriminative power for multi-view faces due to their less
  within-class variance.

  Nonetheless, Haar-like wavelet and its variants are not the only visual descriptor that has gained
  tremendous success, other locally extracted features, \eg,
  edge orientation histograms (EOH) \cite{Levi2004Learning},
  Histogram of Oriented Gradients (HOG) \cite{Dalal2005Histograms},
  Local Binary Pattern (LBP) operator \cite{Ojala2002Multiresolution},
  have also performed remarkably well in vision applications.
  Levi and Weiss \cite{Levi2004Learning} proposed EOH which divides edges
  into a number of bins.
  Three set of features are then used to describe an image region:-
  a ratio between each orientation, a ratio between a single orientation and
  the difference between two symmetric orientations.
  For frontal face detection, EOH achieves state of the art performance
  using only a few hundred training images.
  Dalal and Triggs  proposed histogram of oriented
  gradients in the context of human detection \cite{Dalal2005Histograms}.
  Their method uses a dense grid of histogram of oriented gradients,
  computed over blocks of various sizes.
  Ojala \etal  proposed LBP feature,
  which is derived from a general definition of texture in local neighborhood \cite{Ojala2002Multiresolution}.
  Two most important properties of LBP operators are its invariance against
  illumination changes and its computational simplicity.
  Recently, Wang \etal \cite{Wang2009HOGLBP} combined HOG and LBP descriptors
  as the feature set for human detection.
  The authors reported that the combined classifier yields the best
  descriptor for classifying pedestrians.

  Although these recently proposed descriptors have shown excellent results
  in many empirical studies, when compared to simple Haar-like features,
  they have a {\em much higher complexity} and {\em computation time}.
  Since, face data sets are less complex than human data sets,
  \ie, faces have less variation than human and partial occlusions happen
  less in faces, we simplify the best descriptor reported in Wang \etal \cite{Wang2009HOGLBP},
  namely a combination of HOG and LBP, for the task of face detection.
  Our aim is to lower the time it takes to extract features while
  maintaining their high discriminative power.
  In order to further improve the generalization performance of our simple features,
  we create a more distinctive features combination using sparse least square regression.

%  In this work, we propose two contributions to the work of \cite{Viola2004Robust}.
%  In this work, we introduce simple and efficient edge descriptors which combine
%  the strength of both HOG and LBP descriptors.
%  Although, we only make use of horizontal and vertical edges,
%  our feature lies in a more expressive feature space, which allows us to build
%  classifiers that are much simpler than those obtained with the standard
%  Haar-like wavelet features.
%  Compared to Haar-like wavelet features, our descriptors lie in a more expressive
%  feature space.
%  The second contribution is the use of a Biased Minimax
%  Probability Machine (BMPM) in each cascade layer to specify an upper
%  bound on the probability of misclassification for future data.
%  The BMPM algorithm allows us to exploit the property of asymmetric node
%  learning goal commonly used in cascade framework.

  The rest of the paper is organized as follows.
  Section~\ref{sec:rect} begins by describing the concept of HOG and LBP descriptors.
  We then provide details of our simple edge descriptors, which combine the
  strength of both HOG and LBP descriptors, and propose our joint features.
  Numerous experimental results are presented in Section~\ref{sec:exp}.
  Section~\ref{sec:discussion} provides a brief discussion of our proposed features.
  We conclude the paper in Section~\ref{sec:conclusion}.

\section{Simple Edge Descriptors}
\label{sec:rect}
  Our descriptors are based on HOG and LBP features, which have shown to
  give excellent results in many vision applications.
  The intuition of our descriptors is that the appearance of faces can be
  well characterized by horizontal, vertical and diagonal edges, as shown
  in \cite{Viola2004Robust}.
  Hence, we modify parameters' value used in HOG and LBP to reduce
  the feature extraction time.
  In this section, we first give a brief review of HOG and LBP descriptors.
  We then mention how we adopt HOG and LBP to face detection problem.

  \subsection{HOG and LBP}
  \label{sec:HOGLBP}

  After Lowe proposed Scale Invariant Feature Transformation (SIFT)
  \cite{Lowe2004Distinctive}, many researchers have studied the use of
  orientation histograms in other areas.
  Dalal and Triggs \cite{Dalal2005Histograms} proposed histogram of
  oriented gradients in the context of human detection.
  Their method uses a dense grid of histogram of oriented gradients,
  computed over blocks of various sizes.
  Each block consists of a number of cells.
  A local 1D orientation histogram of gradients is formed from the
  gradient orientations of sample points within a region.
  Each histogram divides the gradient angle range into a predefined
  number of bins.
  The gradient magnitudes vote into the orientation histogram.
  In \cite{Dalal2005Histograms}, each block is quantized into $2 \times 2$ cells
  and the gradient angle in each cell is quantized into $9$ orientations
  (unsigned gradients, \ie, $[0, 180]$ degrees), resulting in a
  $36$-dimensional descriptor ($9$ bins$/$cell $\times$ $4$ cells$/$block).
  In their approach, the final object descriptor is obtained by
  concatenating the orientation histograms over all blocks.

  LBP was first proposed as a gray level invariant texture primitive.
  LBP operator describes each pixel by its relative gray level to its
  neighbouring pixels, \eg, if the gray level of the neighbouring pixel
  is higher or equal, the value is set to one, otherwise to zero.
  Hence, each center pixel can be represented by a binary string.
  The histogram of binary patterns computed over a region is then used
  to describe image texture.
  Fig.~\ref{fig:LBP} illustrates LBP of radius 1 pixel with $8$ neighbours.
  For LBP of radius $1$ pixel, an $8$-bit binary number is generated,
  resulting in $2^8$ distinct values for the binary pattern.
  LBP has several properties that favour its usage, \eg, it is robust
  against illumination changes, has high discriminative power and
  also fast to compute.

  \subsection{Rectangular Features based on HOG and LBP}

\begin{figure*}[tb]
    \begin{center}
    \includegraphics[width=0.7\textwidth,clip]{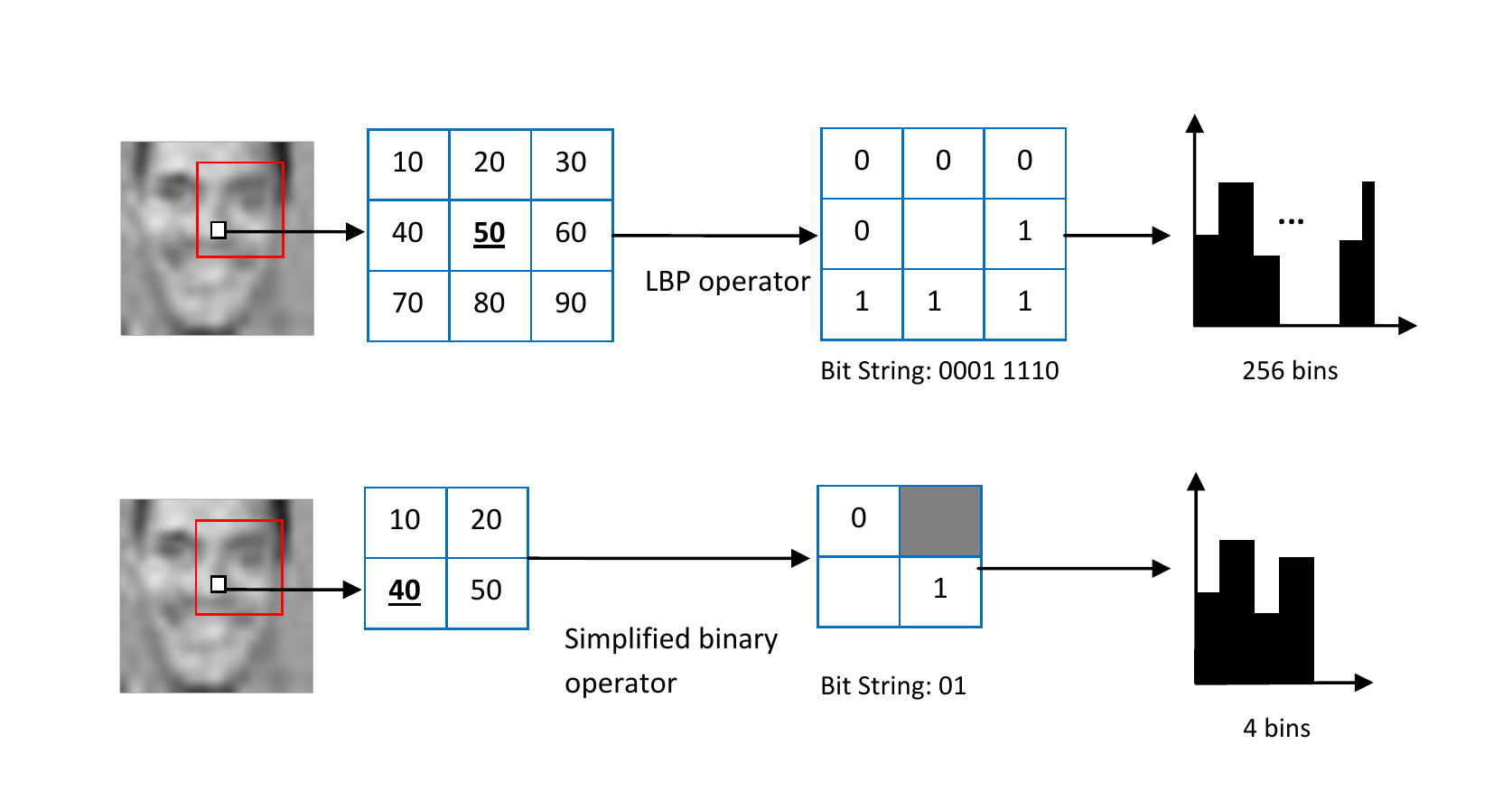}
    \end{center}
    \caption{{\em Top:} An illustration of LBP.
    {\em Bottom:} An illustration of our edge binary pattern.
    Both descriptors have a radius of $1$ pixel.
		}
    \label{fig:LBP}
  \end{figure*}

  Similar to HOG and LBP, we consider the change of pixel intensities in a small image
  neighbourhood to provide a measurement of local gradients inside
  each rectangular region.
  For HOG, we set the number of cells in each block to be one.
  Each block can have various rectangular sizes.
  For LBP, a binary pattern is extracted inside a given rectangular region.
  In this paper, we simplify the computational complexity
  of both HOG and LBP features for fast feature extraction time.
  To achieve this, we quantize the gradient angle into $2$ orientations
  (horizontal and vertical axes).
  We build histogram for both signed and unsigned gradients.
  Hence, each block can be represented by a $4$-D feature vector.
  A vector is normalized to an $ \ell_2$ unit length.
  %Note that by reducing the number of orientation bins to two,
  %HOG is similar to EOH descriptors \cite{Levi2004Learning}.
  We also represent our features similar to LBP by making use of binary
  pattern on a smaller neighbourhood.
  Fig.~\ref{fig:LBP} illustrates our simplified edge binary pattern of
  radius $1$ pixel.
  For each rectangular block, we normalize HOG and LBP separately and
  concatenate them to get the final block descriptor.
  Building on the fundamental concept of HOG and LBP descriptors,
  the new descriptor has many invariance properties such as being tolerance to
  illumination changes, robustness to image noise, and computational simplicity.
  In our paper, multidimensional decision stumps are used as AdaBoost weak learner
  to train our features.

  \subsection{Joint Rectangular Features}

    In our work, we use the AdaBoost classifier with multidimensional decision stumps
    as weak learner.
    One disadvantage of training a weak learner with a single feature
    is that the generalization performance hardly improves in later rounds
    of boosting.
    Many researchers have observed that adding more weak learners
    can reduce the training error but not the
    generalization error \cite{Li2004Float,Wu2008Fast,Mita2008Discriminative}.
    More importantly, the detection performance of single features
    drops drastically in later stages of the cascade.
    We believe that single features are not discriminative
    enough to separate faces from difficult non-faces.

    The use of feature co-occurrences in each weak classifier
    has been shown to yield higher classification performance
    compared to the use of a single feature \cite{Mita2008Discriminative}.
    Similar to Mita \etal \cite{Mita2008Discriminative}, we solve this
    problem by applying the concept of joint features to create a
    more distinctive co-occurrence of features.
    Instead of using class-conditional joint probabilities,
    we approach this problem by using sparse least square regression
    to train our weak classifiers.
    Least square regression is proved to be one of the most effective
    weak classifiers in various literatures
    \cite{Avidan2007Ensemble,Parag2008Boosting}.
    Joint features using sparse least square regression make it
    possible to classify difficult samples that are misclassified by
    weak classifiers using a single feature.

    Let training data sets consist of $n$ samples $(X_i, y_i)$,
    $i = 1..n$, where $X_i = [R_1, R_2, .., R_j]$ is the vector
    of $1$D rectangular features and $y_i$ is an object class
    $\in \{-1, 1\}$.
    The least square model has the form $f(X,\boldsymbol\beta) = \beta_{1} R_1
    + \beta_{2} R_2 + \cdots + \beta_{j} R_j + \beta_{0}$.
    The least square method finds optimum parameters,
    $\hat{\boldsymbol\beta}$, where the weighted sum of
    squared residuals,
    $\sum_{i=1}^{n} { w_i {[ y_i - f(X_{i}, \hat{\boldsymbol\beta}) ]}^2 }$,
    is minimized.
    Here, $w_i$ is the sample weights.
    In order to construct a set of distinctive feature co-occurrences,
    we focus on a subset of rectangular features.
    In other words, we add a sparsity constraint into our least
    square problem.
    The optimization problem can now be defined as
    \begin{align}
      \min_{i} \quad & \sum_{i}
      {w_i {[ y_i - f(X_i, \hat{\boldsymbol\beta}) ]}^2} , \\
      \st        \quad & \card(\boldsymbol\beta) = k, \notag
    \end{align}
    where $\card(\boldsymbol\beta) = k$ is an additional sparsity
    constraint and $\card(\cdot)$ counts the number of nonzero components.
    The problem is non-convex, combinatorial and NP-hard.
    Since, least square problem can be viewed as Generalized Rayleigh
    Quotient problem \cite{Moghaddam2008Sparse}, an efficient
    greedy approach similar
    to the one proposed in \cite{Moghaddam2007Fast} can be adopted here.
    In other words, the optimal solution to sparse generalized
    Eigen-value decomposition \cite{Moghaddam2007Fast} is also the
    optimal solution to our sparse least square regression.

    To improve the generalization performance, a simple decision stump is introduced to each
    rectangular feature.  Hence, each feature value is represented by a decision stump's output
    (binary response), specifying object or non-object, respectively.  The threshold value in
    threshold function is selected based on AdaBoost sample weights in each iteration.

\section{Experiments}
\label{sec:exp}

  This  section is organized as follows.
  The data sets used in this experiment, including how the performance
  is analyzed, are described.
  Experiments and the parameters used are then discussed.
  Finally, experimental results and analysis of different techniques
  are presented.

  \subsection{Frontal Face Detection}
  \label{sec:face}
	
    Due to its efficiency, Haar-like rectangle features
    \cite{Viola2004Robust} have become a popular choice as image features
    in the context of face detection.
    We compare our rectangular features with Haar-like features.
    Similar to the work in \cite{Viola2004Robust}, the weak learning
    algorithm known as decision stumps is used here due to their
    simplicity and efficiency.

    \subsubsection{Performances on Single-node Classifiers}
    \label{fig:single}
      \begin{figure*}[t]
	    \begin{center}
		\subfigure[]{
		  \includegraphics[width=0.45\textwidth,clip]{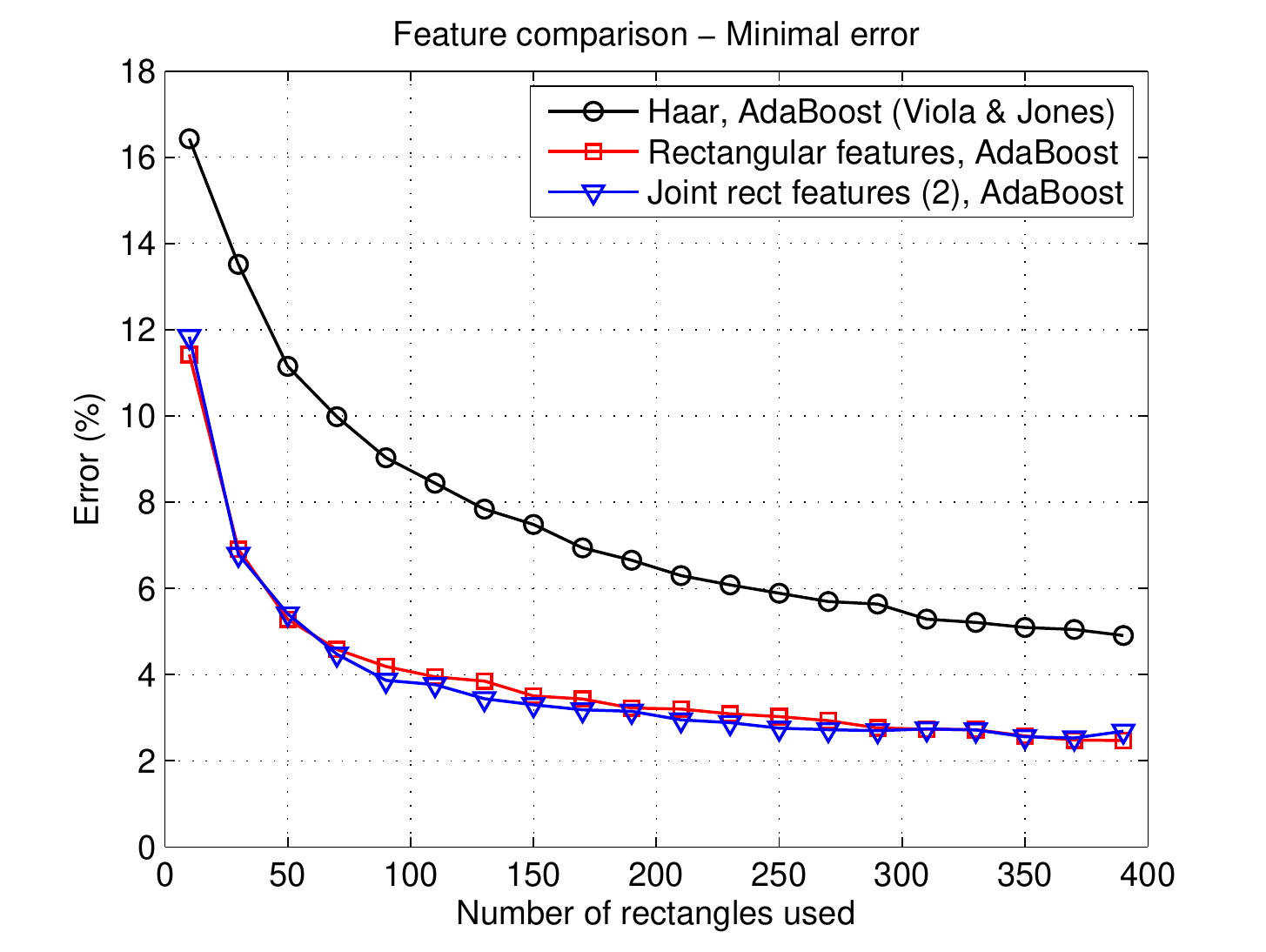}
		  \label{fig:single_a}
		}
		\subfigure[]{
		  \includegraphics[width=0.45\textwidth,clip]{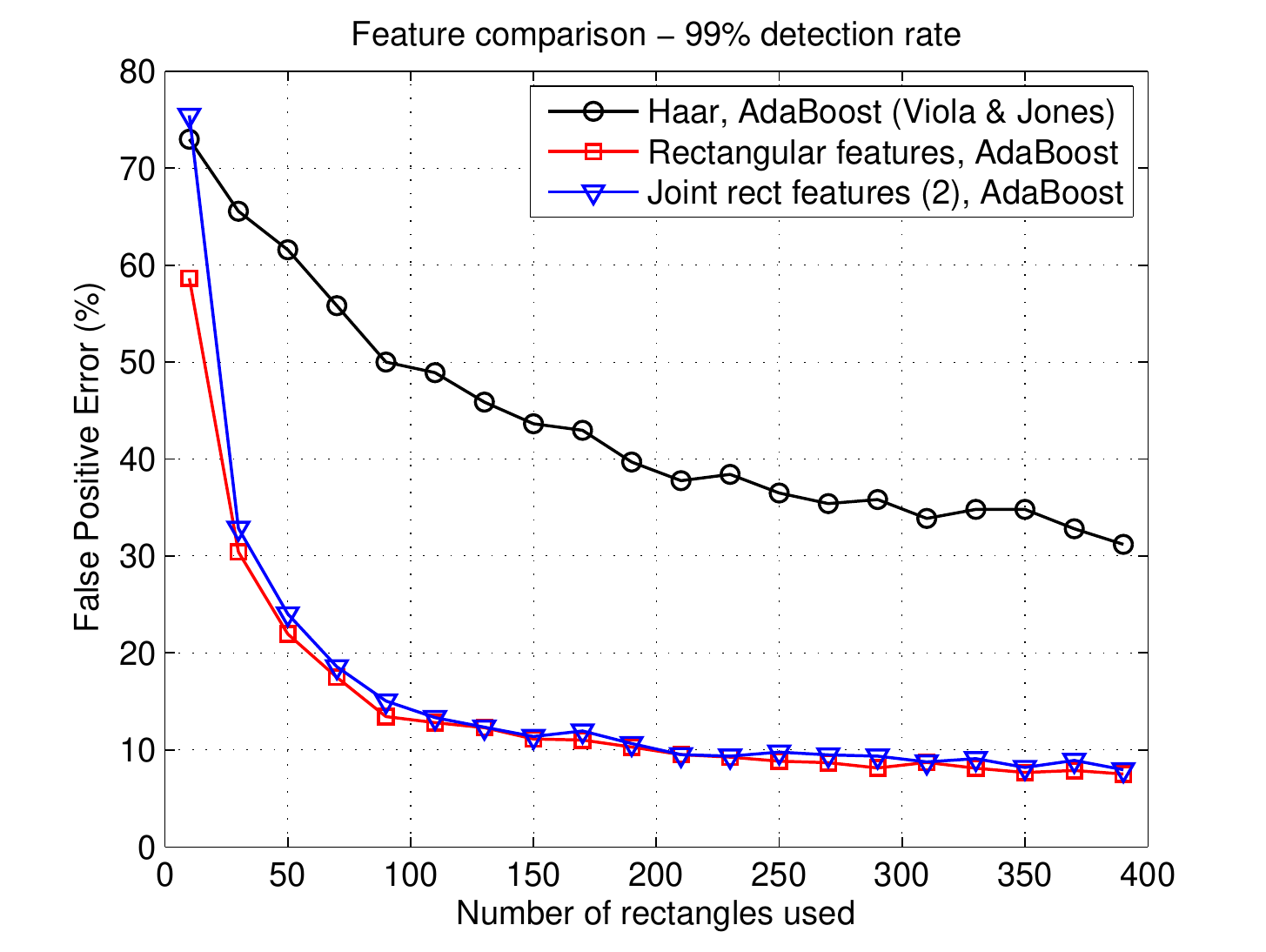}
		  \label{fig:single_b}
		}
		\end{center}
		\caption{ Best viewed in color. Comparison of
		  (a) error rates between Haar-like features
		  and rectangular features. The joint rectangular
		  classifier consists of two rectangles in each
		  weak classifier.
		  (b) false alarm rates on test sets between
		  Haar-like features and rectangular features.
		}
	    %\vspace{-0.25cm}
      \end{figure*}
      %\vspace{-0.05cm}

      In order to demonstrate the performance of our features,
      we replace Haar wavelet like features used in \cite{Viola2004Robust}
      with our features.
      In the first experiment, we compare a single strong classifier
      learned using AdaBoost with Haar wavelet like features
      and our proposed rectangular features.
      The data sets consist of $10,000$ mirrored faces.
      They were divided into three training sets and two test sets.
      Each training set contains $2,000$ face examples and $2,000$
      non-face examples.
      The faces were cropped and rescaled to images of size
      $24 \times 24$ pixels.
      For non-face examples, we randomly selected $5,000$ non-face
      patches from non-faces images and added $5,000$ difficult
      non-faces, for a total of $10,000$ patches.
      % -- I think this experiment is unfair. I should have used
      % -- misclassified faces from other classifier e.g. eigenface, etc.
      %
      %misclassified by \cite{Viola2004Robust} and collected
      %$10,000$ patches in total.
      For each experiment, three different classifiers are generated,
      each by selecting two out of the three training sets and the
      remaining training set for validation.
      The performance is measured by two different curves:-
      the test error rate and the classifier learning goal (the false
      alarm error rate on test sets given that the detection rate on
      the validation set is fixed at $99\%$).

      Experimental results are shown in Figs.~\ref{fig:single_a} and
      \ref{fig:single_b}.
      The following observations can be made from these curves.
      Having the same number of learned features, rectangular
      features achieves lower generalization error rate and
      false positive error than Haar features.
      Based on our observations, Haar features seem to perform
      slightly better than rectangular when the number of features
      is less than $5$.
      This is not surprising since Haar features contain more
      variety of shapes than our rectangular features.
      The first few selected Haar features often combine different
      parts of the faces and therefore would be more discriminative
      than our rectangular shape.
      The performance of our joint rectangular features is also
      shown in the figure.
      On face data sets, we observe a lower error rate when we combine
      two rectangular features.
      Combining three or more rectangular features does not
      improve the performance any further.

	\subsubsection{Performances on Cascades of Strong Classifiers}
	\label{sec:cascade_face}

      \begin{figure*}[tbh]
	    \begin{center}
		  \includegraphics[width=0.45\textwidth,clip]{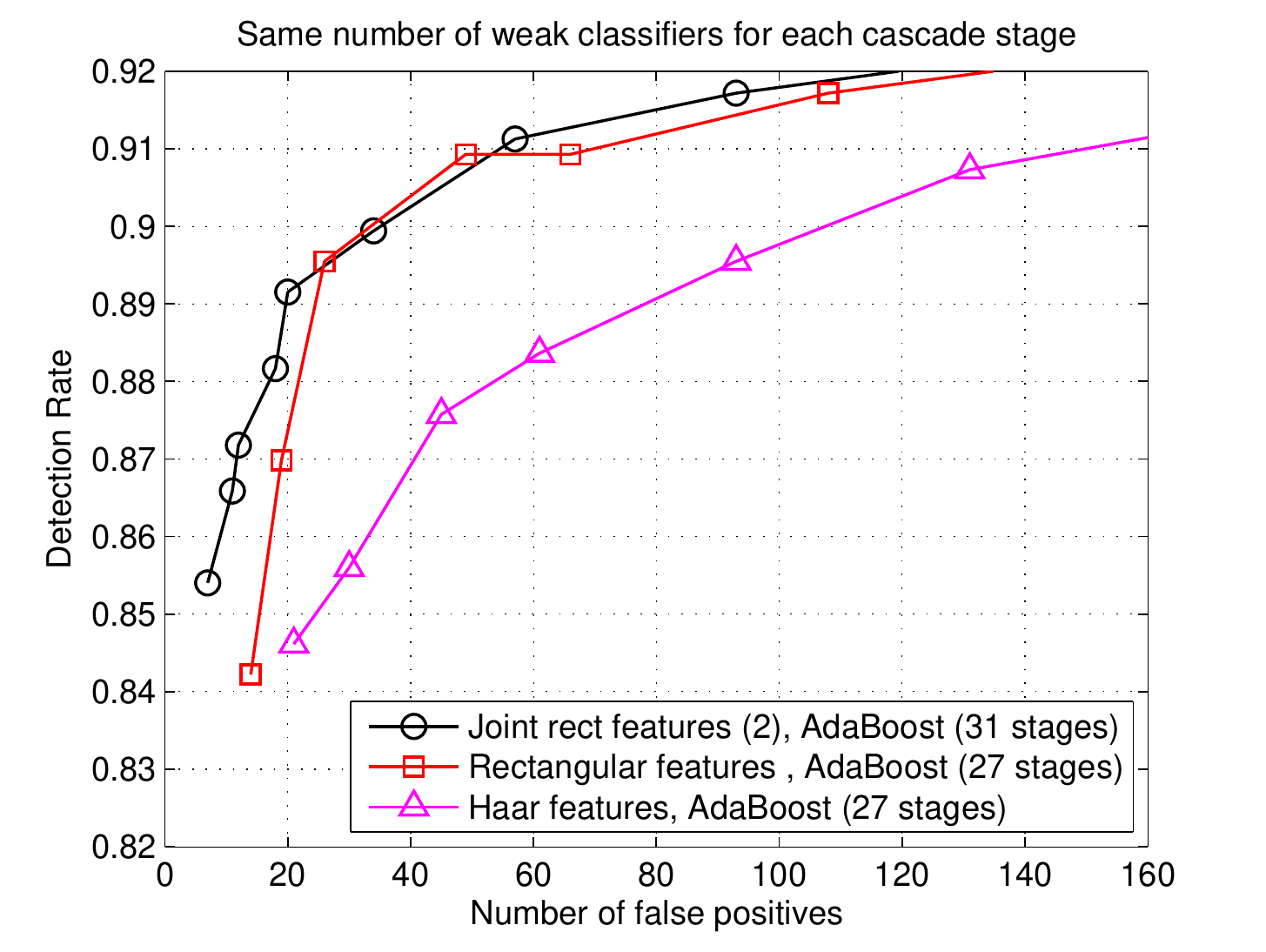}
		\end{center}
		\caption{
        Performance comparison between
		our rectangular features, joint rectangular features
		and Haar features on cascades of strong classifiers.
		}
		\label{fig:cascade_a}
      \end{figure*}

	  In the next experiment, we used $2,500$ frontal faces
	  ($5,000$ mirrored faces) that we obtained from \cite{Viola2004Robust}.
	  All faces were cropped and rescaled to a size of $24 \times 24$ pixels.
	  For non-face examples, we randomly downloaded over $7,000$ images
	  of various sizes from the internet.
	  We used MIT+CMU test sets to test our system.
	  The set contains $130$ images with $507$ frontal view faces.
	  We set the scaling factor to $1.2$ and window shifting step to $1$.
	  The technique used for merging overlapping windows is similar
	  to \cite{Viola2004Robust}.
	  Multiple detections of the same face in an image are considered
	  false detections.

\begin{figure*}[tbh]
	    \begin{center}
		  \includegraphics[width=0.57\textwidth,clip]{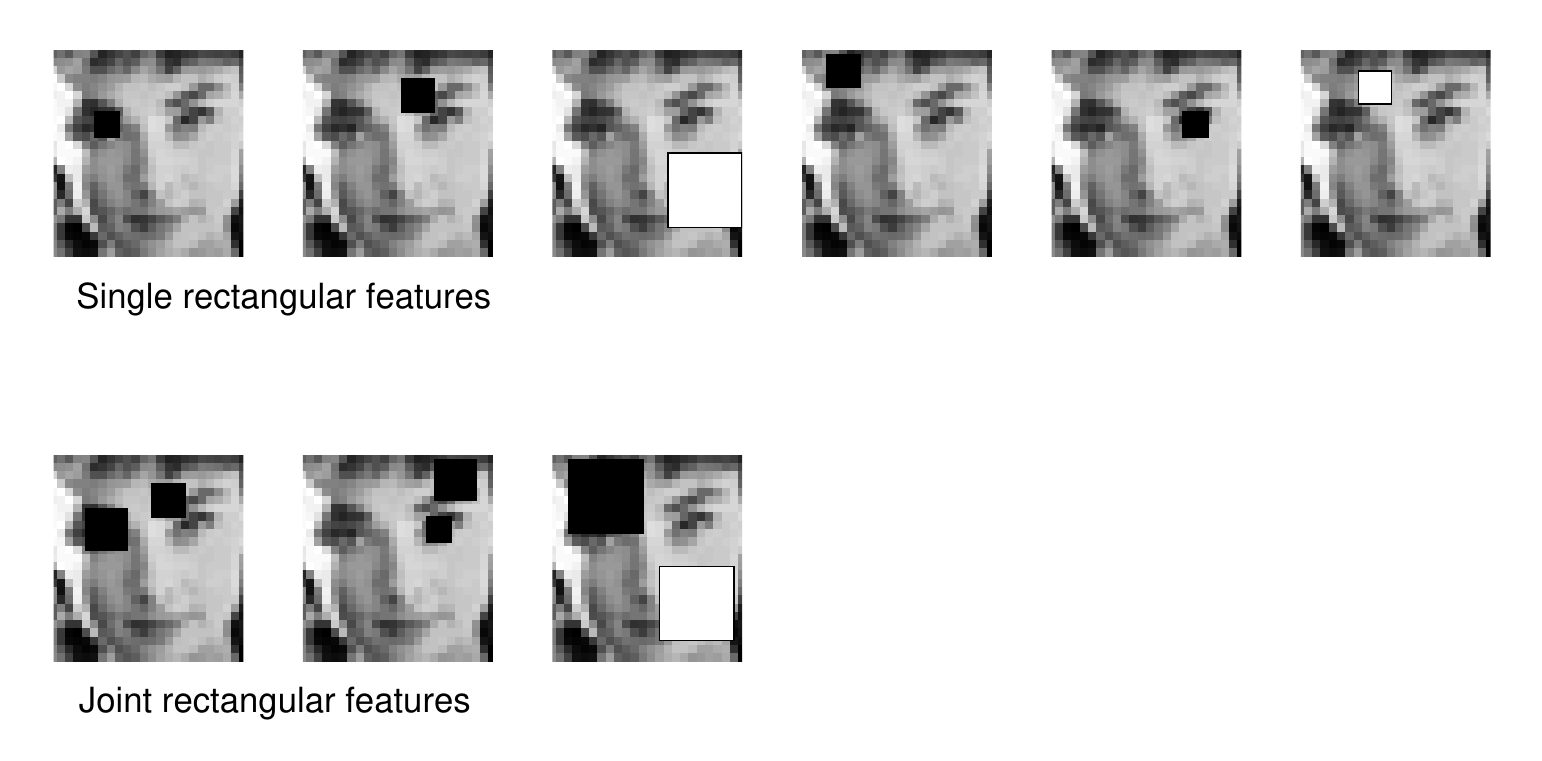}
		\end{center}
		\caption{
        The first few selected rectangular features from the first layer of cascade.
        Black boxes indicate HOG features and white boxes indicate LBP feature.
		}
		\label{fig:rectangle}
      \end{figure*}

      For fair evaluation of both rectangular and Haar-like features,
      we adopted a simple cascade as proposed in \cite{Viola2004Robust}.
      Each cascade layer consists of the same number of features
      (weak classifiers).
      The non-face samples used in each cascade layer are collected from
      false positives of the previous stages of the cascade
      (bootstrapping).
      The cascade training algorithm terminates when there are not
      enough negative samples to bootstrap.
      Fig.~\ref{fig:cascade_a} shows a comparison between the Receiver
      Operating Characteristic (ROC) curves produced by both features.
      The ROC curves show that rectangular features outperform
      Haar-like features at all false positive rates.
      Similar to previous experiment, the combination of two rectangular
      features in each weak classifier performs best.
      From the figure, the performance gap between single and joint features is wider
      at low number of false positives, \ie, at $85\%$ detection rate,
      joint features achieve $10$ less false positives than single features.
      Experimental results indicate that the type of features we use has a
      crucial role in the ability of the system to generalize.
      %
      % Paul 29 Oct, it's probably not an advantage
      %
      %A key advantage of using rectangular features is that the
      %rectangle takes less calculation to compute ($4$ lookup tables)
      %and the resulting classifier has less number of weak
      %classifiers.
      %
      Fig.~\ref{fig:rectangle} shows single and joint rectangular features selected
      in the first cascade layer.
      Most selected patches cover the area around the eyes and forehead.

\begin{figure}[bth]
	    \begin{center}
		\includegraphics[width=0.07\textwidth,clip]{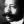}
		\includegraphics[width=0.07\textwidth,clip]{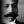}
		\includegraphics[width=0.07\textwidth,clip]{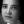}
		\includegraphics[width=0.07\textwidth,clip]{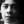}
		\includegraphics[width=0.07\textwidth,clip]{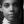}
        
		\includegraphics[width=0.07\textwidth,clip]{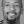}
		\includegraphics[width=0.07\textwidth,clip]{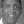}
		\includegraphics[width=0.07\textwidth,clip]{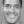}
		\includegraphics[width=0.07\textwidth,clip]{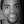}
		\includegraphics[width=0.07\textwidth,clip]{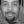}
		
		\includegraphics[width=0.07\textwidth,clip]{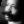}
		\includegraphics[width=0.07\textwidth,clip]{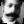}
		\includegraphics[width=0.07\textwidth,clip]{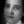}
		\includegraphics[width=0.07\textwidth,clip]{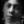}
		\includegraphics[width=0.07\textwidth,clip]{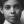}

		\includegraphics[width=0.07\textwidth,clip]{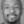}
		\includegraphics[width=0.07\textwidth,clip]{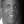}
		\includegraphics[width=0.07\textwidth,clip]{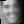}
		\includegraphics[width=0.07\textwidth,clip]{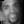}
		\includegraphics[width=0.07\textwidth,clip]{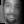}
		
		\includegraphics[width=0.07\textwidth,clip]{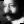}
		\includegraphics[width=0.07\textwidth,clip]{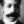}
		\includegraphics[width=0.07\textwidth,clip]{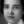}
		\includegraphics[width=0.07\textwidth,clip]{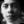}
		\includegraphics[width=0.07\textwidth,clip]{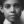}
		
        \includegraphics[width=0.07\textwidth,clip]{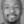}
		\includegraphics[width=0.07\textwidth,clip]{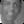}
		\includegraphics[width=0.07\textwidth,clip]{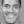}
		\includegraphics[width=0.07\textwidth,clip]{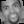}
		\includegraphics[width=0.07\textwidth,clip]{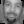}
		
		\includegraphics[width=0.07\textwidth,clip]{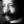}
		\includegraphics[width=0.07\textwidth,clip]{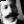}
		\includegraphics[width=0.07\textwidth,clip]{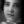}
		\includegraphics[width=0.07\textwidth,clip]{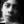}
		\includegraphics[width=0.07\textwidth,clip]{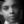}
		
        \includegraphics[width=0.07\textwidth,clip]{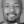}
		\includegraphics[width=0.07\textwidth,clip]{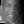}
		\includegraphics[width=0.07\textwidth,clip]{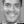}
		\includegraphics[width=0.07\textwidth,clip]{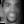}
		\includegraphics[width=0.07\textwidth,clip]{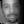}
		\end{center}
		\caption{ Manipulated faces.
		Top row: faces are exposed to random illumination changes and translated randomly.
		Second row: faces are randomly in plane rotated and exposed to
		random illumination changes.
		Third row: faces are randomly in plane rotated and randomly translated.
		Last row: faces are randomly rotated, randomly translated and exposed to
		random illumination changes.
		}
	    \label{fig:manipulated}
      \end{figure}

      Since, face labeling process is rather tedious and time
      consuming; it is quite common that the labeled faces are
      misaligned and rotated.
      In the next experiment, we compare the performance of
      rectangular features and Haar-like features on noisy
      face data sets.
      In other words, we want to determine how much effect
      the noisy training data will have on the detection
      performance.
      We automatically rotate, shift and illuminate faces
      in the training sets using some predefined rules.
      Some of the modified faces are shown in Fig.~\ref{fig:manipulated}.
      Similar to previous experiments, we used AdaBoost to train
      both features.
      Some readers might point out that AdaBoost is vulnerable
      to handling noisy data and the use of other classifiers,
      \eg, LogitBoost \cite{Friedman2000Additive} and BrownBoost
      \cite{Freund2004Adaptive}, would yield better generalization.
      However, this would defeat the purpose of comparing Haar
      features with rectangular features.
      Table~\ref{tab:tab2} shows detection rates of both
      features when trained on different noisy data sets
      and tested on MIT+CMU test sets.
      Based on our results, rectangular features are much
      better at handling noisy training data.
      We notice less performance drop when the classifier is
      trained with rectangular features.

      The disadvantage of rectangular features compared to Haar-like features
      is that we now have to keep $8$ integral images in the memory for
      fast feature extraction (signed and unsigned vertical edge responses,
      signed and unsigned horizontal edge responses and
      $4$ bins for LBP histogram).
      In terms of an evaluation time, rectangular features have a
      higher evaluation time than Haar-like features due to
      an overhead in integral images' calculation.

      \begin{table}[bth]
      \caption{Performance comparison of classifiers using
      rectangular features and Haar-like features on
      noisy data sets.
      Here, we compare the detection rate on MIT+CMU test sets
      when the number of false positives is $50$.
      \textbf{R} means faces are in plane rotated.
      \textbf{L} means faces are exposed to illumination changes (lighting).
      \textbf{M} means faces are translated by a few pixels (misaligned).
      }
      \begin{center}
        \begin{tabular}{l|c|c|c}
        \hline
          $ $    &  Rectangle shaped  &
          Haar-like features & Perf. Improvement\\
        \hline
        \hline
          Original            & $0.905$  & $0.878$  & $3.0\%$ \\
          R+L                 & $0.844$  & $0.790$ & $6.3\%$ \\
          M+L                 & $0.856$  & $0.821$  & $4.3\%$ \\
          R+M                 & $0.835$  & $0.814$ & $2.5\%$ \\
          R+M+L               & $0.826$  & $0.739$ & $12.0\%$ \\
        \hline
          Average             & $0.853$  & $0.808$  & $5.5\%$ \\
        \hline
        \end{tabular}
      \end{center}
      \label{tab:tab2}
      \end{table}

\begin{figure*}[t!]
    \begin{center}
    \includegraphics[width=0.6\textwidth,clip]{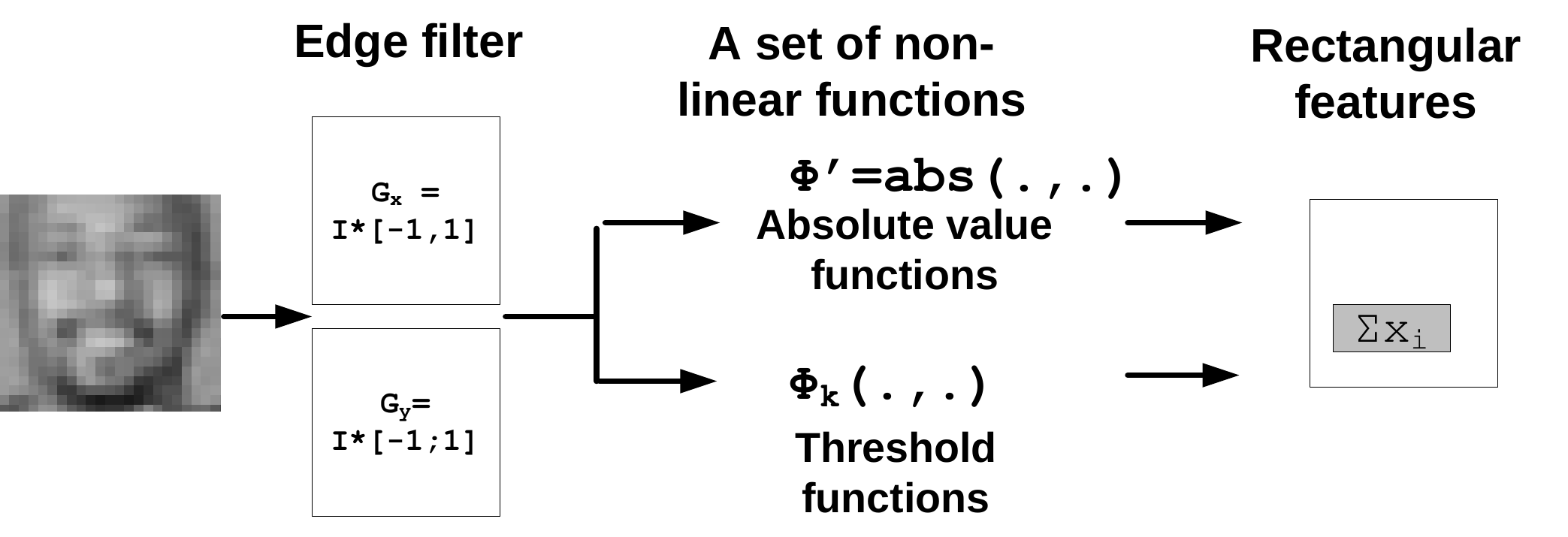}
    \end{center}
    \caption{An illustration of rectangular features.}
		\label{fig:block}
\end{figure*}

\section{Discussion}
\label{sec:discussion}

In this paper, we proposed a simple and robust local feature descriptor for face detection.
Our rectangular features can be denoted by a $4$-tuple, $(x,y,w,h)$, where $x$ and $y$ denote the $x$-coordinate
and $y$-coordinate of the top left position of the block, $w$ and $h$ are the width and height of the rectangles, respectively.
Rectangular features are based on simplified HOG and LBP features.
Our simplified HOG can be viewed as a sum of edge responses, in vertical and horizontal directions.
For unsigned gradients, we apply an absolute value function to edge responses.  The absolute value
of a real number is its numerical value without its sign.  From image processing point of view, the
absolute values of the intensity changes represent the magnitude of the edges without taking into
consideration the polarity of the edges.  Each rectangle is represented by a $4$-D feature vector,
which is normalized to an $ \ell_2$ unit length.  Simplified binary operator can be viewed as
applying the simple threshold function to both vertical and horizontal edge responses.  The
threshold function can be classified as one form of activation functions commonly used in neural
network.  The output of the functions takes on the value of $1$ or $0$ depending on the sign of both
horizontal and vertical gradients, \eqref{EQ:thresh}.

    \begin{equation*}
      \phi_1(x,y) = \left\{ \begin{array}{ll}
                    1 & \mbox{if $x \geq 0$ and $y \geq 0$}; \\
                    0 & \mbox{otherwise},\end{array} \right.
    \end{equation*}
    \begin{equation*}
      \label{EQ:nonlin}
      \phi_2(x,y) = \left\{ \begin{array}{ll}
                    1 & \mbox{if $x \geq 0$ and $y < 0$}; \\
                    0 & \mbox{otherwise},\end{array} \right.
    \end{equation*}
        \begin{equation*}
      \label{EQ:nonlin2}
      \phi_3(x,y) = \left\{ \begin{array}{ll}
                    1 & \mbox{if $x < 0$ and $y \geq 0$}; \\
                    0 & \mbox{otherwise},\end{array} \right.
    \end{equation*}
    \begin{equation}
    \label{EQ:thresh}
      \phi_4(x,y) = \left\{ \begin{array}{ll}
                    1 & \mbox{if $x < 0$ and $y < 0$}; \\
                    0 & \mbox{otherwise}.\end{array} \right.
    \end{equation}
    where $x$ and $y$ denote vertical and horizontal edge responses.

An illustration of our rectangular features based on HOG and LBP is shown in Fig.~\ref{fig:block}.
We can generalize rectangular features as follows.  First, we apply edge filters to the original
image.  Edge filter is one of the most popular techniques used to detect a rate of changes at any
given pixel coordinates.  Edge responses can be calculated from partial derivatives in horizontal
and vertical directions of a given pixel location.  After deriving vertical and horizontal edge
responses, we apply two non-linear functions to these responses; namely absolute value function and
$2$-D threshold function.  By introducing non-linearity into low-level features, we observe an
improvement in the overall performance on visual classification tasks.  These non-linear functions
might look over-simple.  However, many researchers have reported that applying these simple approach
often leads to performance improvement in vision applications, \eg, binary operator has been used in
LBP to describe image texture as described in Section~\ref{sec:HOGLBP}, absolute value function has
also been used in Speeded Up Robust Features (SURF) \cite{Bay2008Speeded} where it performs
remarkably well in describing key-point descriptor.  In summary, our rectangular features consider
the change of pixel intensities in a small image neighborhood to provide an approximate
representation of edge responses inside the specific region.  This finding raises several open
questions related to possible face detection features.  In the future we plan to research on
learning a more efficient rectangular feature, which would be more memory efficient, work well on
general objects and can achieve a comparable speed to Haar-like features.

\section{Conclusion}
\label{sec:conclusion}

In this work, we proposed the use of simple edge descriptors, which combine the
discriminative power of HOG with the strength of LBP operators.
Since, single feature is not discriminative enough to separate faces from difficult non-faces,
we further improve the generalization performance of our simple features by
applying feature co-occurrences.
Experimental results show that our new features not only outperform Haar-like
features but also yield better generalization when training on noisy data.
On average, we achieve a performance improvement of $5.5\%$ when trained with
rectangular features.

\bibliographystyle{IEEEtran}
\bibliography{rectangle}

\end{document}